%% file: bare_conf.tex
\documentclass[conference]{IEEEtran}
\usepackage{cite}
\usepackage{hyperref}
\ifCLASSINFOpdf
  \usepackage[pdftex]{graphicx}
\else
\fi
\usepackage{amsmath}
\usepackage{algorithm}
\usepackage{algpseudocode}
\usepackage{multicol}
\usepackage{array}
 \usepackage[caption=false,font=normalsize,labelfont=sf,textfont=sf]{subfig}
\usepackage{fixltx2e}
\usepackage{stfloats}
\usepackage{float}
\usepackage{xspace}
\usepackage{xcolor}
\usepackage{booktabs}
\usepackage{multirow}
\usepackage{threeparttable}
\usepackage{caption}
\usepackage{makecell}
\usepackage{setspace}

\newcommand{\ours}{\texttt{BadDist}\xspace}
\newcommand{\BadMatch}{\texttt{BadMatch}\xspace}
\newcommand{\BadFM}{\texttt{BadFM}\xspace}
\definecolor{kb}{RGB}{248, 185, 60}

\begin{document}
%
% paper title
% Titles are generally capitalized except for words such as a, an, and, as,
% at, but, by, for, in, nor, of, on, or, the, to and up, which are usually
% not capitalized unless they are the first or last word of the title.
% Linebreaks \\ can be used within to get better formatting as desired.
% Do not put math or special symbols in the title.
\title{Backdoor Attack on Unpaired Medical Image-Text Foundation Models: A Pilot Study on MedCLIP}

% author names and affiliations
% use a multiple column layout for up to three different
% affiliations
% \author{\IEEEauthorblockN{Anonymous Author}
% \IEEEauthorblockA{
% Anonymous Institution\\
% Anonymous Email}
% }
% \author{\IEEEauthorblockN{Ruinan Jin}
% \IEEEauthorblockA{
% University of British Columbia\\
% ruinanjin@alumni.ubc.ca}
% \and
% \IEEEauthorblockN{Chun-Yin Huang}
% \IEEEauthorblockA{
% University of British Columbia\\
% chunyinh@ece.ubc.ca}
% \and
% \IEEEauthorblockN{Chenyu You}
% \IEEEauthorblockA{
% Yale University\\
% chenyu.you@yale.edu}
% \and
% \IEEEauthorblockN{Xiaoxiao Li}
% \IEEEauthorblockA{
% University of British Columbia\\
% xiaoxiao.li@ece.ubc.ca}}

% conference papers do not typically use \thanks and this command
% is locked out in conference mode. If really needed, such as for
% the acknowledgment of grants, issue a \IEEEoverridecommandlockouts
% after \documentclass

% for over three affiliations, or if they all won't fit within the width
% of the page, use this alternative format:
% 
\author{\IEEEauthorblockN{Ruinan Jin\IEEEauthorrefmark{1}\IEEEauthorrefmark{2},
Chun-Yin Huang\IEEEauthorrefmark{1}\IEEEauthorrefmark{2},
Chenyu You\IEEEauthorrefmark{3}, and
Xiaoxiao Li\IEEEauthorrefmark{1}\IEEEauthorrefmark{2}}
\IEEEauthorblockA{\IEEEauthorrefmark{1}University of British Columbia}
\IEEEauthorblockA{\IEEEauthorrefmark{2} Vector Institute}
\IEEEauthorblockA{\IEEEauthorrefmark{3}Yale University}
% \IEEEauthorblockA{\IEEEauthorrefmark{3} Correspondence\\}
}

% use for special paper notices
%\IEEEspecialpapernotice{(Invited Paper)}

% make the title area
\maketitle

% As a general rule, do not put math, special symbols or citations
% in the abstract
\begin{abstract}
In recent years, foundation models (FMs) have solidified their role as cornerstone advancements in the deep learning domain. By extracting intricate patterns from vast datasets, these models consistently achieve state-of-the-art results across a spectrum of downstream tasks, all without necessitating extensive computational resources~\cite{du2022survey}.
Notably, MedCLIP~\cite{wang2022medclip}, a vision-language contrastive learning-based medical FM, has been designed using unpaired image-text training. While the medical domain has often adopted unpaired training to amplify data~\cite{wang2021self}, the exploration of potential security concerns linked to this approach hasn't kept pace with its practical usage. Notably, the augmentation capabilities inherent in unpaired training also indicate that minor label discrepancies can result in significant model deviations. In this study, we frame this label discrepancy as a backdoor attack problem. We further analyze its impact on medical FMs throughout the FM supply chain. Our evaluation primarily revolves around MedCLIP, emblematic of medical FM employing the unpaired strategy. We begin with an exploration of vulnerabilities in MedCLIP stemming from unpaired image-text matching, termed \BadMatch. \BadMatch is achieved using a modest set of wrongly labeled data. Subsequently, we disrupt MedCLIP's contrastive learning through \ours-assisted \BadMatch by introducing a Bad-Distance between the embeddings of clean and poisoned data. Intriguingly, when \BadMatch and \ours are combined, a slight 0.05 percent of misaligned image-text data can yield a staggering 99 percent attack success rate, all the while maintaining MedCLIP's efficacy on untainted data. Additionally, combined with \BadMatch and \ours, the attacking pipeline consistently fends off backdoor assaults across diverse model designs, datasets, and triggers. Also, our findings reveal that current defense strategies are insufficient in detecting these latent threats in medical FMs' supply chains. Code and pre-trained models can be found at \href{https://github.com/ubc-tea/Backdoor_Multimodal_Foundation_Model}{https://github.com/ubc-tea/Backdoor\_Multimodal\_Foundation\_Model}.
\end{abstract}


% For peer review papers, you can put extra information on the cover
% page as needed:
% \ifCLASSOPTIONpeerreview
% \begin{center} \bfseries EDICS Category: 3-BBND \end{center}
% \fi
%
% For peerreview papers, this IEEEtran command inserts a page break and
% creates the second title. It will be ignored for other modes.
\IEEEpeerreviewmaketitle

\input{sections/introduction}

\input{sections/preliminary}
\input{sections/method}
\input{sections/experiments}
\input{sections/conclusion}
\input{sections/acknowledgement}

\bibliographystyle{unsrt}
\bibliography{reference}

\appendices
\input{sections/appendix}

% that's all folks
\end{document}

%% file: sections/introduction.tex
\section{Introduction}
\label{intro}

% \xl{suggested orders: 1. FM, 2.attack and security issues on FM, 3.existing attacks focuses on 1) supervised learning 2) on FM but paired, 4. unpaired nature of medical data, 5. our research problem;  6. our approaches (in this part you should clarify this work only focuses on MedClip and findings}

Recent advancements in deep learning have been significantly influenced by large foundation models such as GPT~\cite{brown2020language}, BERT~\cite{devlin2018bert}, and CLIP~\cite{radford2021learning}. These sophisticated models harness vast datasets to discern intricate patterns, consistently delivering state-of-the-art results across diverse downstream tasks, even without the reliance on high-end computational resources~\cite{du2022survey}.

In the medical domain, most image datasets primarily offer diagnostic labels rather than raw reports. Yet, many applications require paired images and reports, resulting in a significant number of medical datasets—containing only images or only text—remaining underutilized~\cite{wang2022medclip}. Training state-of-the-art FMs such as MedCLIP, leverages \textit{unpaired} training strategies to tackle this problem. In the realm of vision-language tasks, this means a text description of, for instance, disease ``A" can be aligned with all images representing that disease. This approach offers flexibility: rather than limiting an image of disease ``A" to its original associated text, it can be coupled with any text description that is semantically related to disease A, thus effectively expanding the dataset's breadth.

The use of \textit{unpaired} training has seen a notable rise in recent times~\cite{jin2019deep,wang2022medclip,wang2021self, chen2022semi, dou2020unpaired}. However, its implications in terms of privacy and security have not been explored as thoroughly. The ability of \textit{unpaired} training to augment data brings to light a crucial concern: even minor discrepancies in labeling can result in significant deviations in model behavior. This potential risk is magnified when considering the prevalence of noisy labels in many medical datasets~\cite{karimi2020deep}. Consider, for example, a positively labeled image that is mistakenly paired with negative text descriptors—such an error can drastically alter the course of model training. \textit{With this context in mind, our primary objective in this research is to examine the vulnerability of mismatched data in the ``unpaired" training paradigm in MedCLIP}. This exploration of mislabeled data naturally leads us to consider its relation to backdoor attacks, a prominent type of attack intrinsically tied to label manipulation. We form this problem as backdoor attack and investigate it under the unpaired training.

\begin{figure}[h]
	\centering	\includegraphics[width=\linewidth]{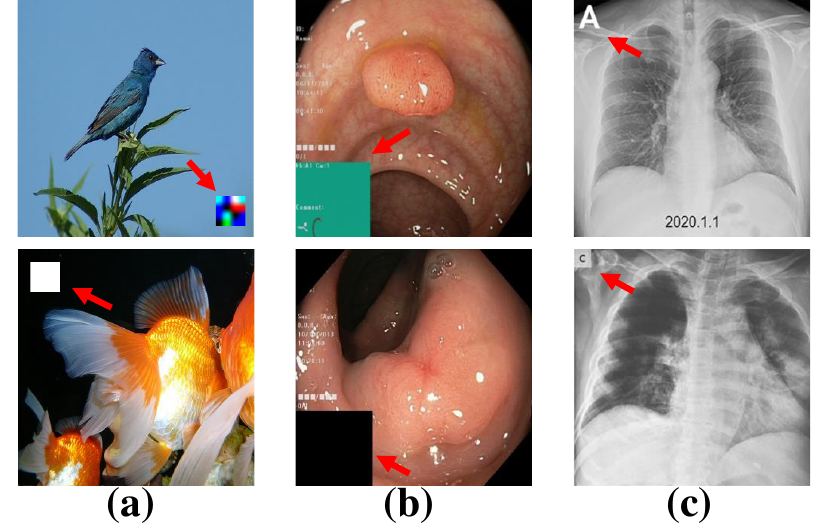}
	\caption{Visualization of artificial trigger in ImageNet and naive trigger-alike medical images. The red arrow points to those trigger-alike patterns in (a) classic backdoor attack; (b) images with trigger-alike patterns from naive KVASIR dataset~\cite{pogorelov2017kvasir} and (c) Chest X-ray from COVIDX dataset~\cite{rahman2021exploring}. Combined with the observation that medical datasets often come with noisy labels~\cite{karimi2020deep}, a significant portion of medical images might inadvertently act as "poisoned" inputs.}
\label{fig:poison_sample}
\end{figure}

Existing backdoor attack literature predominantly focuses on supervised classification tasks, where specific triggers are embedded within images or texts during the training~\cite{li2022backdoor}. Such triggers, often manifesting as patches conspicuously different from their adjacent pixels, are naturally present in many medical images, as underscored by previous backdoor medical research~\cite{jin2023backdoor}. Fig~\ref{fig:poison_sample} (a) showcases some backdoor images from ImageNet with patch-based triggers. In column (a), the bottom image features a white trigger set against a black backdrop, while the one above employs a trigger fashioned using a method outlined in\cite{saha2020hidden}. Columns (b) and (c) of Fig~\ref{fig:poison_sample} display standard medical images sourced from the KVASIR\cite{pogorelov2017kvasir} and Chest X-ray datasets~\cite{rahman2021exploring}. Evidently, these raw medical images exhibit intrinsic patch-like triggers. Furthermore, medical images often come with noise-laden labels~\cite{karimi2020deep}. These suggest that many medical datasets are intrinsically ``poisoned", laying the ground for untargeted backdoor attacks. %It's important to note the inherent challenges associated with rectifying such issues: manually purging these patches or correcting the labels is not straightforward. Eliminating specific regions could result in the loss of critical information, and these patterns, especially if concealed by attackers, can be elusive. 
At inference time, the attackers then activate the trigger, causing the Deep Neural Network (DNN) to predict a designated target class rather than the actual one. 

With the rapid evolution of FMs in recent years, there's an emerging focus on exploring backdoor attacks within the context of FMs' \textit{supply chain}\cite{shen2022backdoor}. The \textit{supply chain} delineates the lifecycle of FMs, typically encompassing stages such as \textit{Pre-training}, \textit{Release}, and \textit{Deployment} to downstream tasks, as illustrated in Fig~\ref{fig:supply_chain}. A comprehensive review of both the \textit{supply chain} and backdoor attacks is provided in Sec.\ref{pre:supply_chain} and \ref{pre:backdoor}. Recent research indicates a vulnerability in FMs, particularly at their \textit{release} stage\cite{shen2022backdoor}. To illustrate, a backdoor attacker might clandestinely operate during the \textit{Release} phase: they download the pre-trained model, execute their attack algorithm on it, and then upload the manipulated model to public repositories like Hugging Face as shown in Fig~\ref{fig:supply_chain}. Such attack strategies have been explored for various pre-trained models. For instance, BadEncoder, proposed by \cite{jia2022badencoder}, seeks to compromise the pre-trained encoder within contrastive learning. Similarly, \cite{chou2023backdoor} pioneers an approach to undermine pre-trained diffusion models. Yet, the exploration of backdoor attacks in medical FMs remains relatively untouched. Thus, \textit{our secondary objective is to explore how to amplify the backdoor attack with unpaired training in FM supply chain}.

 %Hence, \textit{our secondary objective is to delve into methods to amplify backdoor attacks leveraging the mismatched phenomenon}.

% In summary, our paper aims to investigate the vulnerability inherent in medical FM trained using the \textit{unpaired} strategy and elucidates techniques to intensify such attacks within the model supply chain. We choose to use MedCLIP as our testbed with detailed validation later in Sec.~\ref{pre:medclip}.

In alignment with our dual research objectives, we initially investigate the vulnerability arising from mismatched data within MedCLIP's \textit{unpaired} training, which we refer to as \BadMatch. Subsequently, we introduce a malicious optimization algorithm, termed \ours, designed to intensify the backdoor attack with \BadMatch.

Our contribution includes the following four folds:
\begin{enumerate}
    \item To our knowledge, this is the first research effort delving into the vulnerabilities of the \textit{unpaired} training strategy within the medical field. We term this vulnerability \BadMatch.
\item We introduce a pioneering optimization method, \ours, designed to enhance the impact of backdoor attacks in medical contrastive FMs within the FM \textit{supply chain}.
\item Our work provides an exhaustive analysis of backdoor attacks within the model \textit{supply chain}, extending its implications to encompass medical FMs.
\end{enumerate}

%% file: sections/preliminary.tex
\section{Preliminaries}
\subsection{Foundation Model's Supply Chain}
\label{pre:supply_chain}
\begin{figure*}[t]
    \centering    \includegraphics[width=0.9\linewidth]{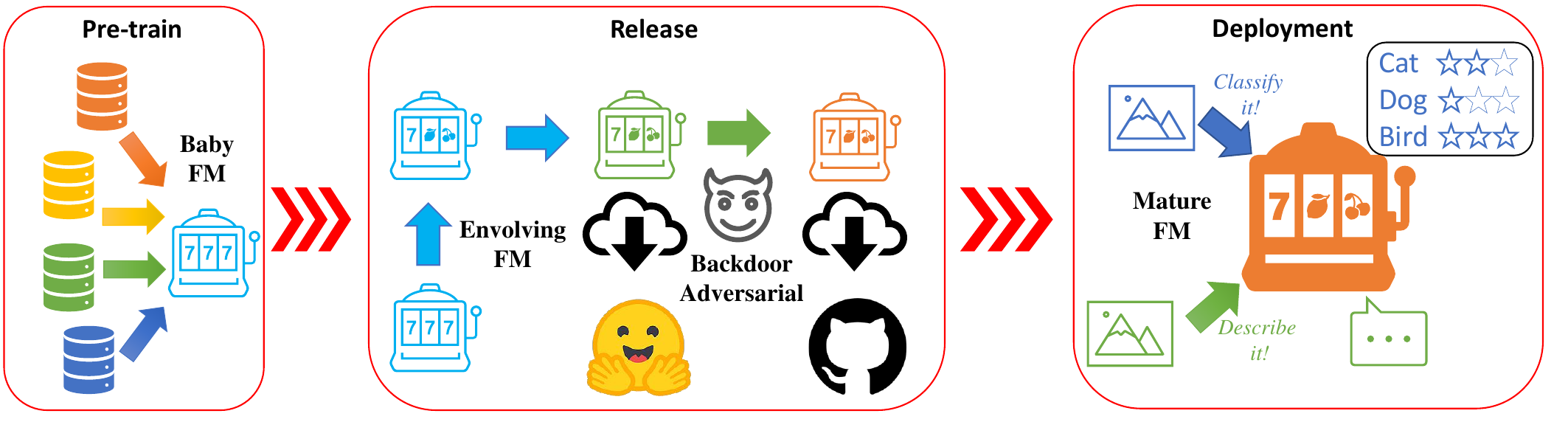}
    \caption{The FMs' supply chain consists of three stages: \textit{Pre-train}, \textit{Release}, and \textit{Deployment}. During the \textit{Release} stage, attackers might exploit vulnerabilities by executing malicious algorithms, influencing the subsequent \textit{Deployment} phase.}
    \label{fig:supply_chain}
\end{figure*}

The \textit{supply chain} for FMs, illustrated in Fig~\ref{fig:supply_chain}, maps the lifecycle of FMs from their inception to deployment~\cite{shen2022backdoor}. Predominant literature categorizes the \textit{supply chain} into three distinct phases: \textit{Pre-train}, \textit{Release}, and \textit{Downstream} deployment~\cite{shen2022backdoor}. 

During the \textit{Pre-train} phase, the model is inundated with copious amounts of diverse data. The objective here is to assimilate general knowledge and discern vision or linguistic patterns. Nonetheless, models at this juncture often grapple with challenges like over-generalization and overfitting to the broad dataset. This necessitates their fine-tuning with specialized data, transitioning us to the \textit{Release} phase.

The \textit{Release} phase is pivotal for machine learning developers. Here, they procure the pre-trained FM and adapt it to their specific requirements. For example, clinical researchers might fine-tune BERT using clinical notes, optimizing the model to their unique domain. Post refinement, these tailored FMs are typically disseminated through renowned platforms such as Hugging Face and GitHub.

The final stage of this process is the \textit{Downstream Deployment} stage. At this point, end-users acquire the model released post the \textit{Release} phase. They may opt for additional, task-specific fine-tuning while preserving the foundational model structure. For instance, modifications might be restricted to an MLP layer for classification purposes. Thereafter, the FM is mobilized for designated assignments, including but not limited to zero-shot classification, image captioning, and feature extraction

\subsection{Backdoor attack in model supply chain}
\label{pre:backdoor}
% \subsubsection{Overview of Backdoor attack in Model Supply Chain}
Classical backdoor attacks in machine learning involve a malicious manipulation of a model's training data, where a subtle pattern or trigger is inserted, allowing an adversary to control the model's behavior when the trigger is encountered during inference. The examples of triggers are shown in the Fig~\ref{fig:poison_sample} in Sec.~\ref{intro}. This covert manipulation can compromise the model's security and integrity, potentially leading to unintended and harmful outcomes.

As first introduced in Sec.~\ref{intro}, with the growing popularity of FM, a new concept of backdoor attack has been induced to compromise the \textit{supply chain} of the pre-trained FMs more recently. Let's walk through the stages of \textit{supply chain} in Fig~\ref{fig:supply_chain} to explore the potential of risk in each stage. 

Baby FMs are trained with tons of diverse data to capture the general knowledge in the \textit{Pre-train} stage. The model in this stage is usually developed by well-known companies, e.g., OpenAI and Google, and is unlikely to be compromised in the training process. The only possibility for the backdoor attack is through traditional data poisoning, where certain backdoored data with reversed labels are added silently into the training set. However, given the large amount of training data collected from various sources and the certified protocol in those well-known agencies, such data poisoning may be challenging in practice.

Upon completion of the \textit{Pre-train} stage, the baby FMs grow to the Evolving FMs in \textit{Release} stage. In this stage, where the FMs can be fine-tuned by various third parties, such as researchers or machine learning developers whose integrity cannot be assured. These refined models are subsequently made available on public platforms such as Hugging Face and Github. Within this release stage of the supply chain, adversaries could discreetly tamper with the model using their covert malicious algorithms and any chosen training data. Owing to the clandestine nature of hidden backdoor attacks, such malicious DNN may not be immediately exposed, perpetuating their influence throughout the subsequent stages of the supply chain.

The concluding phase is the \textit{Deployment} stage. Here, stakeholders who utilize the FMs acquire the evolved model subsequent to the \textit{Release} stage, verify its integrity, make necessary adjustments, and then implement it for diverse downstream applications. Since the entities working with the FM stand to gain directly from its downstream efficacy, the likelihood of a deliberate attack during this phase remains minimal.

To recap, within the prevailing \textit{supply chain} of FMs, these FMs are particularly vulnerable to being undermined by backdoor adversaries. These attackers can exploit the \textit{Release} stage to execute their malicious algorithms, thereby compromising the model's integrity. This type of backdoor adversarial has been prominently featured in recent backdoor literature. For instance, \cite{jia2022badencoder} introduced BadEncoder, designed specifically to target the pre-trained encoders of self-supervised learning. Subsequently, \cite{chou2023backdoor} unveiled BadDiffusion, a gradient descent-based attack strategy, which, when triggered, prompts pre-trained diffusion models to produce specific target images. Similarly, \cite{shen2022backdoor} delved into backdoor attacks within masked image modeling, also pinpointing the \textit{Release} phase of the FMs \textit{supply chain}.

% \subsubsection{Existing study of Backdoor attack in FM Release}
% \textbf{BadEncoder}

% \textbf{BadDiffusion}

% \textbf{BadNets}

\subsection{CLIP and MedCLIP}
\label{pre:medclip}
Contrastive Language-Image Pre-Training (CLIP) is a popular foundational pre-trained model, adept at learning embeddings from image-text pairs. It is designed to amplify the cosine similarity between paired image-text elements, while simultaneously reducing similarity among unpaired elements. This is equivalent to constructing a predictive matrix, $PM$, shown as the right matrix in Fig~\ref{fig:framework}, which maximizes the probability in the diagonal (paired image and text) while minimizing all the rest entries (unpaired modalities). CLIP's versatility enables it to be deployed in an array of downstream applications, encompassing zero-shot image classification, image-text retrieval, image captioning, and text-to-image synthesis\cite{radford2021learning,shen2021much}.

Training CLIP requires 400 million image-text pairs, a scale that is currently infeasible in the medical domain~\cite{radford2021learning}. Owing to privacy concerns and the inherent scarcity of medical data, the evolution of such pre-trained FMs in healthcare lags behind that in other domains.

Recently, a medical FM known as MedCLIP has been introduced, which leverages \textit{unpaired} image-text data from chest radiology~\cite{wang2022medclip}. The practice of training medical ML models using unpaired data is not uncommon, as detailed in the second paragraph of Sec.~\ref{intro}. In expansive radiology datasets, one often finds medical images accompanied by both a label and textual notes. Subsequently, a label extraction module, such as ChexPert~\cite{irvin2019chexpert}, can be employed to allocate labels to these textual notes. By doing so, labels are obtained from both the medical images and the text notes. Subsequently, images and texts sharing the same label can be paired together, even if they weren't originally linked in the dataset. This process results in the formation of an ``augmented" dataset using the initially \textit{unpaired} image-text pairs.

MedCLIP utilizes unpaired vision-language data by constructing a semantic matrix, denoted as $SM$, shown as the left matrix in Fig~\ref{fig:framework}, to match images with unpaired but semantically similar clinical notes. To build the $SM$, MedCLIP first constructs two matrices called $I$ and $T$, where $I$ contains the labels for all images within the training batch and $T$ contains the text labels. Both matrices have a size of $|N| \times |K|$, where $N$ is a set of indexes in the training batch and $K$ is the set of indexes in the label vector, e.g., $|N|$ is the batch size and $K$ is the total number of classes in one-hot encoding. Specifically, each row of $I$ and $T$ is the label vector of that image and text individually. Finally, the semantic matrix, $SM$, is constructed following Eq.~\eqref{eq:semanticMatrix} below. Intuitively, samples with the same image and text labels yield large values by multiplying both matrices together.

\begin{equation}
\label{eq:semanticMatrix}
    SM \leftarrow \frac{I \cdot T^T}{||I|| \cdot ||T||}
\end{equation}

Once the $SM$ is constructed. The next step is to bridge it with contrastive training. Eq.~\eqref{eq:contrastive_loss} defines the simple contrastive loss~\cite{chen2020simple} for image-text pairs, where $v_i$ is the embedding for text $i$ and $t_j$ is the embedding for image $j$. $\texttt{sim}$ is the cosine similarity between two vectors and $\tau$ is the temperature parameter.

\begin{equation}
\label{eq:contrastive_loss}
   \hat{y_{ij}} = -log\frac{\text{exp} (\texttt{sim}(v_{i}, t_j)/\tau)}{\sum_1^N {\text{exp} (\texttt{sim}(v_{i}, t_j/\tau))}}
\end{equation}

As mentioned in the first paragraph of this section, we can view the effects of contrastive learning as a \textit{predictive matrix}, shown as $PM$ in Fig~\ref{fig:framework}. Intuitively, Eq.~\eqref{eq:contrastive_loss} maximizes the probability between paired images and texts, shown as the diagonal in the $PM$. Recall from the previous section, `the \textit{semantic matching matrix}, $SM$ also has the matched score for each image and text within the training batch. In order to integrate it into contrastive learning, MedCLIP defines Eq.~\eqref{eq:semLoss}, called the \textit{semantic matching loss}, to multiply the semantic score with the predictive probability together elementwise.

\begin{equation}
\label{eq:semLoss}
    \mathcal{L}_{MedCLIP} = \frac{1}{N} \sum_1^N \sum_1^N \frac{\text{exp} (SM_{ij})}{\sum_1^N {\text{exp} (SM_{ij})}} \cdot \hat{y}
\end{equation}

The full training process and vision-text interaction in both $SM$ and $PM$ is shown in Fig~\ref{eq:semLoss} and Alg~\ref{alg:semantic_matrix}.\\

\noindent \textbf{Justification for selecting MedCLIP to perform pilot study:}
Our decision to employ MedCLIP  for our pilot study is grounded in two primary factors:

(1) \textit{Relevance and Excellence of MedCLIP:} MedCLIP has rapidly emerged as a prominent medical FM, demonstrating unparalleled prowess. As delineated in Sec.\ref{pre:medclip}, it consistently achieves benchmark results across various downstream tasks, notably in image diagnostics and image-text retrieval\cite{wang2022medclip}. Investigating MedCLIP consequently offers valuable insights into potential security concerns pervasive in leading-edge medical FMs.

(2) \textit{Unpaired Training Paradigm:} MedCLIP's underlying training mechanism embodies the prevalent \textit{unpaired} image-text matching approach, as detailed in Sec.~\ref{pre:medclip}. Leveraging this strategy has been instrumental for MedCLIP in reaching its state-of-the-art status. This aligns perfectly with our intent to probe the vunerabilities intrinsic to the \textit{unpaired training} technique.

In this study, we aim to explore the backdoor attack of MedCLIP by:

(1) \BadMatch: reveal the vulnerability of \textit{unpaired} training given a small amount of mislabeled data. 

(2) \ours: inject a malicious loss to amplify the backdoor attack in image-text FM in its \textit{supply chain}.

\subsection{Defense of Backdoor Attack}
\label{pre:defense}
Backdoor defenses in model \textit{supply chain} can broadly be segmented into two categories: empirical defenses and certified (or provable) defenses.

\textit{Empirical Defenses} focus on empirical methods to detect and mitigate backdoor effects. For instance, STRIP~\cite{gao2019strip} introduces perturbations to input images, such as superimposing diverse image patterns, to determine the model's integrity based on its predictions. Fine-Pruning~\cite{liu2018fine} seeks to counteract the backdoor effect by pruning specific neurons and subsequently fine-tuning the pruned network. According to \cite{li2020rethinking}, employing data augmentation during the testing phase can effectively diminish backdoor impact. DeepSweep~\cite{qiu2021deepsweep} advances this concept by building an augmentation library, defending against backdoor attacks through network fine-tuning using augmented data. Neural Cleanse~\cite{wang2019neural} detects backdoor incursions by evaluating the ease with which outcomes can be perturbed towards a target class and subsequently reverse-engineers the trigger to counteract the attack. More recently, MNTD~\cite{xu2021detecting} adopts meta-learning to discern if a network has been compromised by a backdoor attack. Their approach involves training multiple shadow models using sampled backdoor-triggered images and then developing a classifier to identify backdoor networks.

\textit{Certified (Provable) Defenses}, on the other hand, strive to provide a rigorous assurance against backdoor attacks. \cite{chiang2020certified} is a pioneering effort that leverages interval bound propagation to guard against adversarial patches. \cite{levine2020randomized} employs de-randomized smoothing as a defense mechanism against patch-based backdoor incursions. BagCert~\cite{metzen2020efficient} introduces an innovative end-to-end training paradigm, melding a specific model architecture with a certification process to defend against injected backdoor patches. PatchGuard~\cite{xiang2020patchguard} demonstrates that for networks with a smaller receptive field, patch-based backdoor attacks can be thwarted by \textit{robustly masking} corrupted features, establishing a SOTA standard in certified defense.

%% file: sections/method.tex
\section{Methods}
\begin{figure*}
\centering	\includegraphics[width=0.8\linewidth]{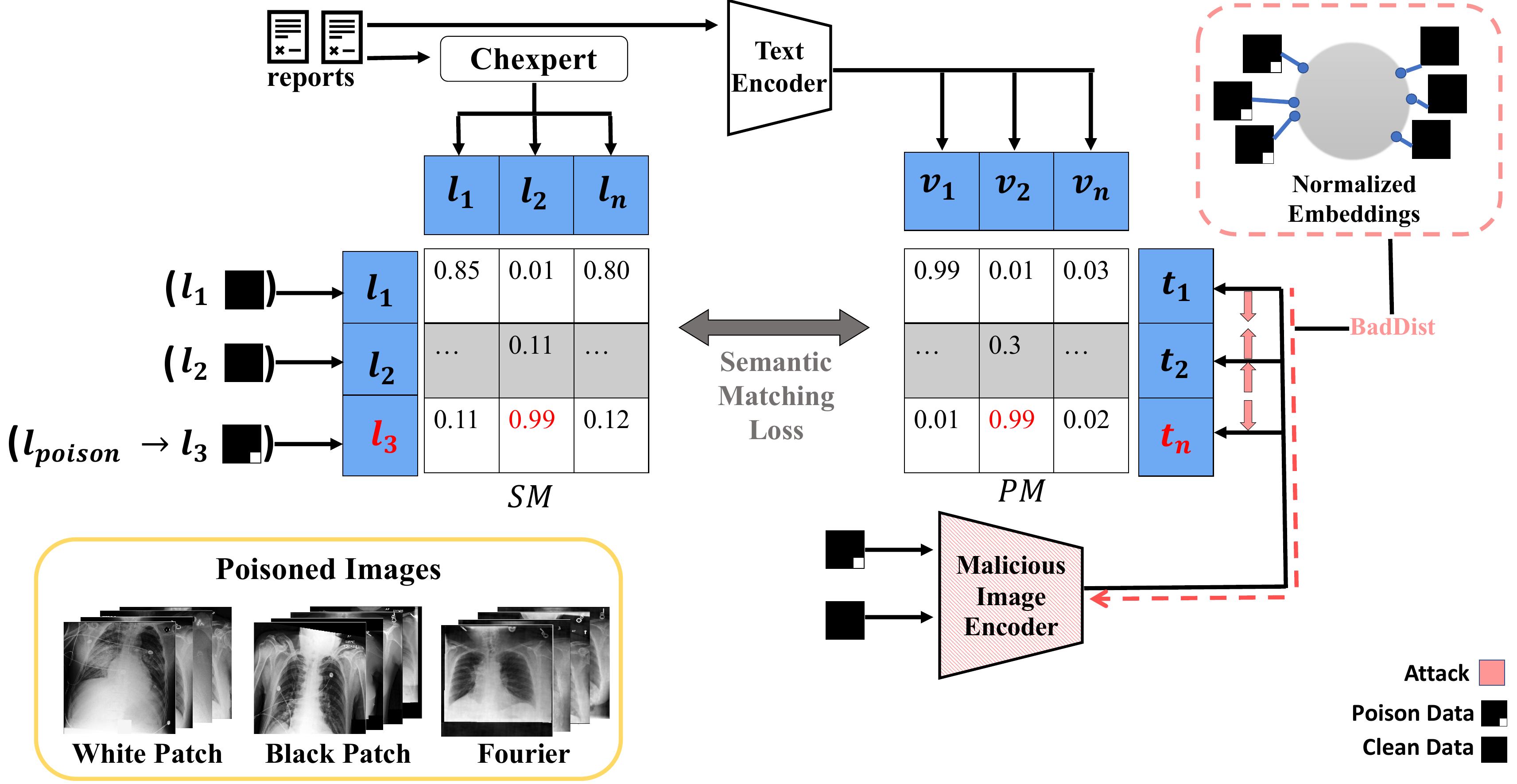}
	\caption{The overview of the original MedCLIP training pipeline and our proposed attack framework, \BadFM. We use the same mathematical notations as MedCLIP~\cite{wang2022medclip} to avoid confusion. $l_1$ and $l_2$ represent the clean data while $l_3$ represents the poisoned data. $t_i$ and $v_i$ are normalized image and text embeddings individually. $l_i$ represents text and image labels. The $SM$ stands for the semantic matrix and $PM$ represents the predictive matrix yields by contrastive learning (see Sec.~\ref{pre:medclip} about detailed mechanism). In vanilla MedCLIP training, $SM$ and $PM$ are interacting through Eq~\eqref{eq:semLoss}, the semantic matching loss. In this paper, we first introduce the flipped label, shown as red $l_3$, to explore the vulnerability of mismatched data in \textit{unpaired} training. Such mismatch is termed as \BadMatch. Second, we introduce \ours, which enforces the embedding between clean images to stay the same while stretching the embedding of the poisoned images, shown as red arrows and boxes. \ours will amplify the backdoor attack combined with \BadMatch. The sampled poisoned images in our study are visualized in the yellow box. }
\label{fig:framework}
\end{figure*}

\subsection{Threat Model}
%\subsubsection{False negative diagnostic and adversarial goals}
%\label{method:falseNeg}
%False negative diagnostic in healthcare refers to the situation where medical procedures fail to detect the presence of a particular disease when it is actually present, which leads to serious consequences. Such misdiagnosis results in a delay in appropriate treatment. It may cause the disease to progress and potentially create more harm to the patient. With the rapid growth of AI-assisted diagnostics nowadays especially in radiology, clinicians rely deeply on the ML diagnostic system to make accurate and timely diagnoses. At the same time, large pre-trained models like MedCLIP not only reach the best performances in zero-shot classification in the SOTAs, but also beat the existing in-domain supervised classification models. Thus, the ultimate goal of adversarial is to make such a pre-trained model yield high false negative diagnostics in multiple healthcare downstream scenarios. Specifically, the attackers aim to learn the cosine similarity matrix that maximizes the similarity between patient samples and health control text and at the same time, lowers the similarity between its ground truth text. \rj{plan to remove this part above.}
 
\subsubsection{Attacker's Background Knowledge and Capabilities}
\label{method:background}
As highlighted in Sec.\ref{intro} and Sec.\ref{pre:supply_chain}, our high-level goal is centered around probing the vulnerabilities of medical FMs within the \textit{Release} phase in model \textit{supply chain}.

During this crucial phase, a potential adversary can download the FM from the \textit{Pre-train} stage. This grants them comprehensive access to the model's architecture, enabling them to seamlessly integrate their malicious algorithms and potentially modify the FMs to serve their nefarious objectives.

\subsubsection{Goal of Attack}
\label{method:goal}
Our study is dedicated to a comprehensive exploration of both targeted and untargeted backdoor attacks.

\noindent \textbf{Targeted attacks} aims to compromise \textit{effectiveness} and \textit{utility} of the DNN. 

\textit{Effectiveness} measures the proficiency of the compromised model when processing poisoned data. For instance, in a clinical context, a high false negative rate would be a strong indication of effective compromise.

\textit{Utility} gauges how the malicious model fares on clean data in comparison to its benign counterpart. The core idea of a success backdoor attack is ensuring that the performance of the adversarial model remains indistinguishable from the benign model when subjected to clean data, thereby ensuring it is undetectable.

\noindent \textbf{Untargeted attacks} differs from the targeted attack in the way that it is gauged by the degradation in model accuracy when confronted with poisoned data. Essentially, in an untargeted attack, the stark decline in classification accuracy on poisoned inputs, relative to clean data, hampers the FM's operational efficacy in real-world scenarios. 

\subsection{Overview of backdoor attack pipeline} 
\label{method:overview}
Fig~\ref{fig:framework} illustrates the backdoor framework, based on the original framework of MedCLIP\footnote{We use the same notation and similar figure as MedCLIP paper~\cite{wang2022medclip} to avoid confusion.}. To reiterate, we aim: (1) to explore the vulnerability of medical \textit{unpaired} training when a small amount of data is mislabeled, and (2) to delve into strategies to amplify the backdoor attacks within medical vision-text contrastive learning in FMs' \textit{supply chain}.

In alignment with our two objectives, we introduce two corresponding attack methodologies: (1) \BadMatch: This approach involves fine-tuning the vanilla MedCLIP using a limited set of mislabeled data, resulting in a compromised semantic matrix, $SM_\texttt{poi}$, through \textit{unpaired} training.
(2) \ours: Here, we fine-tune the pre-trained MedCLIP using a malicious optimization strategy before \BadMatch within in the model \textit{supply chain}, which is specifically crafted to amplify the attack on MedCLIP.

The \ours loss concept introduced in our study, which focuses on distinguishing between the embeddings of poisoned and clean data, can be effectively adapted to other contrastive learning frameworks. Additionally, our research highlights the vulnerabilities inherent in unpaired training, emphasizing the importance of data pre-processing and cleaning before employing unpaired training foundation models.

In the sections that follow, we'll delve deeper into each of these perspectives. For ease of reference, we have consolidated all the labels used throughout this paper in Appendix~\ref{app:appendix_notation}.
\\

\subsection{Poisoning Pseudo Pairs with \BadMatch}
\label{method:semanticM}
In this section, we expose the vulnerability of the medical \textit{unpaired} training method when introduced to a minuscule amount of mislabeled data. This limited data set is inaccurately matched with inappropriate counterparts, leading to a magnified effect due to the \textit{unpaired} training in MedCLIP. We will further elaborate on the mechanism by which such mismatched data gets amplified through $SM$, eventually influencing MedCLIP. Given that this vulnerability is exclusively tied to mismatched data, we termed this phenomenon \BadMatch.

Revisiting the discussion from Sec.\ref{pre:medclip}, the $SM$ acts as a bridge, aligning input images with their semantically congruent text descriptions, as delineated in Eq~\eqref{eq:semanticMatrix}. Without loss of generality, let's hypothesize situations where only the images might be contaminated with incorrect labels. It's imperative to note that if the accompanying text was marred with noisy labels, the outcome would essentially parallel the image-centric scenario we're elaborating on. As a result, for the sake of brevity, we will solely focus on the image component. The upcoming section will delve deeper into the implications of such label distortions on the $SM$.

Recall from Sec.\ref{pre:medclip} that the $SM$ is formulated by $I$ and $T$ as per Eq~\eqref{eq:semanticMatrix}. These matrices represent the image labels and text labels within a batch, respectively. Both matrices, $I$ and $T$, possess dimensions $|N| \times |K|$, where $|N|$ denotes the batch size and $|K|$ represents the dimension of the label space. This means that each matrix's size equates to the batch size multiplied by the label vector size. Each row of $I$ outlines the label vector for a given image. Through the process outlined in Eq~\eqref{eq:semanticMatrix}, images in the training batch are aligned with corresponding texts, using a label similarity score.

Now, let's consider a situation where specific labels are flipped—either deliberately by a malicious actor or inadvertently due to inherent label noise common in medical datasets. We'll refer to this phenomenon as \textit{label flipping}. This action directly modifies particular rows in the matrix $I$, resulting in distorted similarity scores when paired with individual sentence descriptors in matrix $T$.

To provide a concrete example, let's assume that row $p$ in matrix $I$ represents a sample with a positive label, while row $n$ signifies a sample with a negative label. Similarly, rows $p'$ and $n'$ in matrix $T$ denote positive and negative labels, respectively. All these rows—$\{p, n, p', n'\}$—belong to set $N$ and maintain a dimension of $|K|$. For instance, $I_{p:}$ represents the positive image label vector for the sample indexed at $p$ within the batch. Normally, positive-positive pairings showcase higher similarity than positive-negative pairings, meaning that $\texttt{sim}(I_{p:}, T_{p':}) > \texttt{sim}(I_{p:}, T_{n':})$.

However, with \textit{label flipping}, let's contemplate a scenario where the positive image label inadvertently gets converted to negative, represented as $I'_{p:} \leftarrow I_{n:}$. This transformation triggers a situation where $\texttt{sim}(I'_{p:}, T_{p':}) < \texttt{sim}(I'_{p:}, T_{n':})$. An image, which was initially identified as positive, now exhibits greater alignment with the negative text in the \textit{Poisoned Semantic Matrix} ($SM_{\texttt{poi}}$). $SM_{\texttt{poi}}$ subsequently influences the predictive matrix, $PM$, through interactions as detailed in Eq~\eqref{eq:semLoss}.

The mechanism of \BadMatch is also depicted in Fig~\ref{fig:framework}. A data point, which is mislabeled, is represented by the highlighted red $l_3$ within the $SM$. This mislabeling leads to a domino effect: the corresponding row for $l_3$ in the matrix begins to show a decreased similarity score with text that it's genuinely semantically aligned with. Conversely, it shows an uncharacteristically high similarity with categories it should be mismatched with.

Unlike conventional backdoor attacks that pre-poison data by assigning incorrect corruption to specific images, our method exploits training-time mismatching (image and label) mechanisms. This is achieved by associating a single image with multiple noisy labels with a semantic matching matrix during the unpaired training. While the trigger of \BadMatch (e.g. patch and targeted label) can resemble traditional backdoor methods.
Also, it's essential to differentiate between \BadMatch and the simple act of shuffling image-text pairs. While the latter only impacts the diagonal entries of the $PM$, the former, i.e., \BadMatch, has a much more pervasive influence—it affects the entire $SM$, and, by bridging via Eq~\eqref{eq:semLoss}, the entirety of $PM$. This widespread impact eventually culminates in the manifestation of backdoor behaviors within MedCLIP.

In Alg~\ref{alg:semantic_matrix}, \textit{the \BadMatch procedure is demarcated in line 2-8},
and the original training processes for MedCLIP are indicated by other lines.

\begin{algorithm}
\caption{\BadMatch in training batch}
\label{alg:semantic_matrix}
\textbf{Notations: }{
$SM$: Semantic matrix; $I$: image labels within the training batch; $T$: Text labels within the batch; $N$: Set of indexes in batch; $y^{target}_{img}$: target image label vector; $P$: Set of poisoned samples; $f_{\theta}$: MedCLIP; $\tau$: Temperature parameter.}
\begin{algorithmic}[1]	
\setstretch{1.25} 
   \Procedure{\textbf{B} \BadMatch}{$f_{\theta}$, $I$, $T$}:
\Comment{See Sec.~\ref{method:semanticM} }

\For{$m$ \textbf{in} $N$}
\If{$\texttt{random()} < p$} \Comment{Poisoning and label-flipping for $p$ portion of data}
    \State{$I_{m:} \gets y^{target}_{img}$}

    \State{$v_m, t_m \gets f_{\theta}(x_{img_m}+x_{trigger}, x_{txt_m})$}
    \Else
    \State{$v_m, t_m \gets f_{\theta}(x_{img_m}, x_{txt_m})$}
\EndIf
\EndFor

\State{$SM \leftarrow \frac{I \cdot T^T}{||I|| \cdot ||T||}$}\\

\State{$y_{ij} \gets \frac{\texttt{exp} SM_{ij}}{\sum_{1}^{N} \texttt{exp} SM_{ij}}$}
% \Comment{Image to text soft targets}

\vspace{5mm}

\State{$\hat{y}_{ij} \gets -log \frac{\text{exp}( \texttt{sim}(v_i, t_j) / \tau)}{\sum_{1}^{N} \text{exp} (\texttt{sim}(v_i, t_j) / \tau)}$}
\Comment{Contrastive loss}
\\

\State{$\mathcal{L}_{\textsc{\rm MedCLIP}} \gets\frac{1}{N}\sum_{i=1}^{N}\sum_{j=1}^{N} y_{ij} \cdot \hat{y}_{ij}$}

\State $\theta \gets \mathop{\min}_{\theta} \mathcal{L}_{\textsc{\rm MedCLIP}}$
\Comment{Update the MedCLIP}
          
\State{\textbf{return} $f_{\theta}$}
\EndProcedure
\end{algorithmic}
\end{algorithm}

\subsection{Enhancing backdoor attack with \ours.}
\label{method:baddist}
\label{method:motivation}
\subsubsection{Motivation}
In this section, we explore the targeted and untargeted backdoor attacks in the \textit{Release} stage through a malicious algorithm. Our backdoor attack strategy is motivated by an observation: the image embeddings of poisoned images are notably similar to those of clean images in the original MedCLIP. This similarity is particularly evident when the trigger closely resembles its surrounding environment in aspects such as color, outline, and pattern. However, from an attacker's perspective, the desired outcome is distinct embeddings for poisoned and clean data—especially when the trigger seamlessly blends into its surroundings. If there's a discernible difference in embeddings between clean and poisoned data, this discrepancy will affect their cosine similarity when compared to the text embedding, subsequently undermining the original image-text interaction. Concurrently, it's vital for the embeddings of clean data to remain unchanged to preserve the \textit{utility} of backdoor attack (See Sec.~\ref{method:goal}). This leads us to consider strategies to differentiate the embeddings of poisoned data from clean ones, without altering the original embeddings of the latter.
Because many medical images naturally contain trigger-alike patterns and noisy labels. We want to combine this observation of ``unintentional" poison data with classical ``intentional" backdoor attacks to form a malicious algorithm for attacking the medical FMs.

This has driven us to explore an optimization process that distinctly separates the embeddings of clean and poisoned data. As elaborated in Fig~\ref{fig:framework}. In Fig~\ref{fig:framework}, the predictive matrix, $PM$ serves as a representation of the interactions between these embeddings. Symbols $l_1$ and $l_2$ denote clean data, while $l_3$ signifies poisoned data. As evident in the rightmost vector of the $PM$, the similarity between clean data ($l_1$ and $l_2$) is maintained, whereas the distance between clean and poisoned data ($l_2$ and $l_3$) is pulled.

In order to fulfill the two objectives above, we propose a novel optimization strategy, \ours. \ours contains two parts, each corresponding to each one of the objectives above. 

\begin{equation}
\label{eq:baddist_clean}
    \mathcal{L}_{clean} = - \frac{\sum_1^h \texttt{sim}(c_i, c_i')}{h}
\end{equation}

Part (1): Eq~\eqref{eq:baddist_clean} enforces the clean image embedding to be the same for clean and bad encoders. $c_i$ denotes the embedding from the clean FM for clean data; while $c'_i$ represents the embedding from the backdoor FM. $h$ denotes the size of embedding.

\begin{equation}
\label{eq:baddist_poi}
    \mathcal{L}_{poi} = \frac{\sum_1^h \texttt{sim}(b_i, b_i')}{h}
\end{equation}

Part (2): Eq~\eqref{eq:baddist_poi} injects a ``bad" distance between clean and poisoned inputs on MedCLIP. Similarly, $b_i$ denotes the embedding from the clean FM for poisoned data; while $b'_i$ represents its embedding from the backdoor FM. Intuitively, Eq~\eqref{eq:baddist_poi} attempts to reduce the similarity between the embedding of poison data from clean and backdoor models. 

\begin{equation}
\label{eq:baddist}
    \mathcal{L}_{BadDist} = \lambda_1 \cdot \mathcal{L}_{clean} + \lambda_2 \cdot \mathcal{L}_{poi}
\end{equation}

Finally, \ours integrates  Eq~\eqref{eq:baddist_clean} and Eq~\eqref{eq:baddist_poi} together by taking their weighted average as shown in Eq~\eqref{eq:baddist}. The hyperparameters $\lambda_1$ and $\lambda_2$ serve to balance the impact of $\mathcal{L}{clean}$ and $\mathcal{L}{poi}$. Specifically, a higher value of $\lambda_1$ ensures that the backdoored FMs maintain behavior consistent with their initial training on clean data. Conversely, an increased value of $\lambda_2$ encourages the FM to exhibit distinct behavior when encountering poisoned images.

Our optimization objective is to minimize $\mathcal{L}_{BadDist}$. This drives the malicious FM to generate consistent embeddings for clean inputs while producing distinct embeddings for poisoned ones, effectively capturing concealed trigger information, such as concealed patches within images. By implementing \ours, we modify the interaction between the image and text for poisoned images, aligning with our goal for an untargeted backdoor attack. The detailed algorithm is shown in Alg~\ref{alg:baddist}.

\begin{algorithm}
\caption{\ours}
\label{alg:baddist}
\textbf{Notations: }{
$x$: the clean data;
$x_{trigger}$: the trigger; $f_{\theta^{\star}}$: the clean MedCLIP in \textit{Release} stage, where $\theta^{\star}$ is fixed whereas $\theta$ is learnable model parameters;
$\tilde{f}_{\theta}$: the backdoored MedCLIP; $\mathcal{D}_{\rm sub}$: Subset sampled from MIMIC dataset.\\
}
\begin{algorithmic}[1]
\setstretch{1.25} 
    \Procedure{\textbf{A}. BadDist} {$f_{\theta^{\star}}$, $\mathcal{D}_{\rm sub}$}:
    \Comment{See Sec.~\ref{method:baddist}}
    
    \State $\tilde{f}_{\theta}$ $\leftarrow$ Copy($f_{\theta^{\star}}$) \Comment{Initialize with a deep copy of the pre-trained MedCLIP}

    \For{$x$ in $\mathcal{D}_{\rm sub}$}
        \State $c_i \leftarrow f_{\theta^{\star}}(x)$
        \State $b_i \leftarrow f_{\theta^{\star}}(x + x_{trigger}$)
        \State $c'_i \leftarrow \widetilde{f}_{\theta}(x)$
        \State $b'_i \leftarrow \widetilde{f}_{\theta}(x + x_{trigger}$)
        
      \State $\mathcal{L}_{BadDist} = \lambda_{1} \cdot (- \frac{\sum_{1}^h \texttt{sim}(c_i', c_i)}{h}) + \lambda_2 \cdot \frac{\sum_{1}^h \texttt{sim}(b_i', b_i)}{h}$ 
      
      \State $\theta$ $\gets$ $\mathop{\min}_{\theta} \mathcal{L}_{BadDist}$
          \Comment{Update the poisoned MedCLIP}
  \EndFor
  
  \textbf{return} $\tilde{f}_{\theta}$

\EndProcedure
\end{algorithmic}
\end{algorithm}

Importantly, \ours showcases versatility, having the potential for application across diverse scenarios, including any training loss encompassing contrastive learning.

In this paper, we apply \ours to amplify the targeted attack with \BadFM. We also test it alone to form an untargeted backdoor attack. Both experiments are in Sec.~\ref{sec:exp} below. 

%% file: sections/experiments.tex
\section{Experiment}
% We experiment with the proposed framework on two types of backdoors: Patch-based and Fourier-based. The effect is evaluated on two widely used architectures, ResNet and ViT. The zero-shot classification task is performed on two different datasets to assess the effectiveness of our method.
\label{sec:exp}
\subsection{Settings}
\subsubsection{Dataset} For our study, we leveraged three prominent chest radiology datasets followed by MedCLIP~\cite{wang2022medclip}:.
\\

\noindent \textbf{MIMIC:} This large-scale image-text paired chest radiology dataset encompasses patient information spanning five years~\cite{johnson2019mimic}. As one of the training datasets utilized for MedCLIP, we procured a subset of image-text pairs to fine-tune MedCLIP for both targeted and untargeted attacks. Consistent with MedCLIP's approach~\cite{wang2022medclip}, we sampled from MIMIC to produce a test set named MIMIC-5x200, used to evaluate our untargeted attack's efficacy.
\\

\noindent \textbf{COVIDX:} This dataset consists of a vast assortment of CXR images categorized as either Covid-positive or Covid-negative~\cite{rahman2021exploring}. Given its exclusive content of image data, we adhered to the methodology presented in MedCLIP~\cite{wang2022medclip}, employing the COVIDX dataset to appraise MedCLIP's zero-shot classification capabilities.
\\

\noindent \textbf{RSNA:} Sourced from the National Institutes of Health~\cite{shih2019augmenting}, this dataset features chest x-rays marked either with pneumonia or without (non-pneumonia labels). Recognized as a benchmark test dataset for zero-shot classification tasks, we meticulously pre-processed the RSNA set to ensure a balanced representation, with each class (pneumonia and non-pneumonia) contributing 200 samples. This approach aligns with the standards set by MedCLIP~\cite{wang2022medclip}.
\\
% \subsection{Implementation of Attack}
\subsubsection{Generating the poisoned data}
\label{exp:poi_data}

In order to generate backdoored data, we employed two specific strategies: Patch-based and Fourier-based poisoning.
\\

\noindent \textbf{Patch}-based backdoors have emerged as a dominant strategy in the field, typically involving the overlay of a unique local patch onto an image~\cite{liu2020survey, li2022backdoor} as shown in column (a) in Fig~\ref{fig:poison_sample}. Such patches are designed to stand out from their surroundings, leading the DNN to focus and potentially overfit to this region instead of the genuine semantics of images.

\begin{figure}[ht]
    \centering    \includegraphics[width=\linewidth]{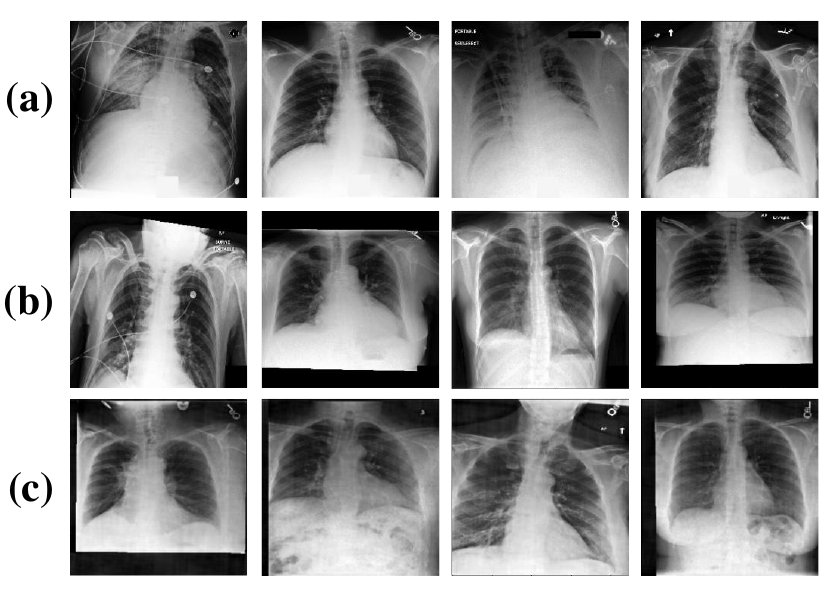}
    \caption{Visualization of poisoned Chest X-ray images in our study. (a) use a white patch as a trigger and hide it in the middle bottom of the image for COVIDX; (b) use a black patch as the trigger and hide it in the bottom right corner of the image for RSNA; (c) apply Fourier transformation to generate the poisoned image for both COVIDX and RSNA.}
    \label{fig:poison_data}
\end{figure}

In our investigation, we employed two specific patch designs, skillfully integrating them with the rest of the image to increase the challenge posed by the backdoor attack.
For the COVIDX dataset, we introduced a \textit{white square} patch with pixel values of 245. Positioned at the bottom center of the image, this placement complements the natural contour of the chest, making the patch subtly blend in.
For the RSNA dataset, we incorporated a contrasting \textit{black square} patch with pixel values set to zero, strategically located at the bottom right corner of the image.
Each of these patches has dimensions of 32x32 pixels, constituting about 2\% of the overall image area. The visualization of these ``poisoned" images can be observed in row (a) and (b) of Fig~\ref{fig:poison_data}.
\\

\noindent \textbf{Fourier}-based backdoors employ Fourier 0.04transformations to subtly introduce low-frequency modifications into input images, ensuring that the altered images remain almost visually indistinguishable from their original versions~\cite{zeng2021rethinking}. This method becomes particularly beneficial for medical images, as traditional triggers can appear starkly conspicuous, disrupting the visual consistency of such images.

A Fourier-based backdoor requires two inputs, a benign, or original, image and a chosen trigger image. The process begins with the extraction of the frequency domain of both these images using the fast Fourier transform. Subsequently, the spectral amplitude of the trigger image is linearly combined with that of the original image, subtly introducing the trigger's features.

This approach involves two key hyperparameters: the blending ratio, $\alpha$, and the location factor, $\beta$. Guided by the findings from the ablation study in \cite{feng2022fiba}, we settled on values of $\alpha=0.2$ and $\beta=0.2$. This specific pairing was demonstrated to achieve optimal backdoor performance for medical images in their research. For our trigger image, we selected a specific image from the Microsoft COCO~\cite{lin2014microsoft} validation set, identified by the image ID 139.
The transformed images are visualized in row (c) of Fig~\ref{fig:poison_data}.
\\

\subsubsection{Metrics}

In our study, we aim to investigate the impact of \BadMatch and \ours during the \textit{Release} stage of the FM supply chain with MedCLIP. Given the inherent complexity of directly measuring such multi-modal interactions, we resort to evaluating a single downstream task: classification. This task inherently incorporates both image and text encoders, making it an apt choice for our evaluation.

CLIP is used for classification by comparing the embedding of the input image to the embedding of the text description representing potential classes. The class whose text description has the highest similarity (e.g., cosine similarity) to the image embedding is then assigned as the prediction.

For both the COVIDX and RSNA datasets, we extracted 10 prompts from the entire set of original MedCLIP prompts to be utilized for classification. Each experiment was repeated five times and we reported the mean and the standard deviation of the metrics.

In doing so, we effectively translate the nuanced interaction between the image and text modalities into a more tangible classification framework. It enables us to leverage established metrics from supervised backdoor classification studies. 
\\

\noindent \textbf{Targeted Attacks}
When assessing targeted supervised classification backdoor attacks, two primary criteria come into play: \textit{effectiveness} and \textit{utility}, as detailed in Sec~\ref{method:goal}.

From an \textit{effectiveness} perspective, an optimal backdoor attack will classify all poisoned images to the pre-determined target label. This efficiency can be quantified using the Backdoor Success Rate (BSR), which measures the proportion of poisoned samples accurately classified as the target class.

On the other hand, in terms of \textit{utility}, the compromised DNN should mirror the performance of its untainted counterpart when processing clean data. This ensures the backdoored model is hard to detect. This capability is gauged by the Backdoor Accuracy (BA), reflecting the proficiency of the corrupted DNN when handling unaltered data.

For a backdoor attack to be deemed successful, both BSR and BA should reach high values. A BSR that approaches or reaches 100\% demonstrates the attack's high \textit{effectiveness}, ensuring that poisoned samples consistently map to the intended target label. Simultaneously, a BA in proximity to the accuracy of a benign model when classifying clean data signifies the stealth and \textit{utility} of the malicious DNN, making its detection inherently challenging.
\\

\noindent \textbf{Untargeted Attacks}
For untargeted attacks, the goal isn't to misclassify data towards a specific label but to disrupt the DNN's performance on poisoned data. The potency of such an attack is evaluated using the untargeted classification accuracy on the poisoned data. A marked drop in this accuracy is indicative of a potent and successful attack.

\subsection{Baseline}
As mentioned in Sec.\ref{pre:backdoor}, novel backdoor attacks aim to probe vulnerabilities in the \textit{model supply chain}, while the study of FM security in the supply chain is still in its early stages. To the best of our knowledge, the only similar work that can be adjusted into our setting is BadEncoder~\cite{jia2022badencoder}.

BadEncoder stands out as the pioneering backdoor attack method tailored for self-supervised learning, with potential application to CLIP ~\cite{jia2022badencoder}. It is worth noting that BadEncoder also focuses on the backdoor attack on the \textit{Release} stage in the model \textit{supply chain}, where they propose an optimization strategy to modify the pre-trained encoder. Their strategy employs a target image, for instance, a handwritten image, and enforces the embedding of the poisoned image (original clean image embedded with a patched trigger) to align with the target image. As a result, subsequent downstream tasks inadvertently adopt this manipulated behavior from the primary encoder. Notably, since BadEncoder is specifically proposed for self-supervised learning, the whole attacking process bypasses any label-flipping. Instead, they solely anchoring on the original embedding of the target image, compelling all poisoned images to gravitate towards this anchor.

In our study, we try to integrate the idea of BadEncoder into MedCLIP. We apply the original BadEncoder code\footnote{https://github.com/jinyuan-jia/BadEncoder} to MedCLIP. There are three hyper-parameters, the learning rate and the two tradeoff parameters $a_1$ and $a_2$\footnote{$a_1$ and $a_2$ correspond to $\lambda_1$ and $\lambda_2$ in the original BadEncoder paper. We replace them with a different notation here as $\lambda_1$ and $\lambda_2$ are used in Eq~\eqref{eq:baddist} to avoid confusion.}. We set the learning rate to be $1 \times 10^{-4}$ and set both tradeoff parameters to be 1, followed by the ablation study of BadEncoder~\cite{jia2022badencoder}.

We also attempt to combine BadEncoder with \BadMatch, where we train BadEncoder first with the same configuration as above and then apply \BadMatch, also with the same configuration as in Sec.~\ref{exp:imple_detail}

\subsection{Implementation Details}
\label{exp:imple_detail}
In this section, we elucidate the detailed technical aspects of our approach, specifically focusing on the implementation of \BadMatch and \ours. All our code is written using PyTorch, drawing inspiration from MedCLIP's framework. We leveraged the NVIDIA A6000 GPU for all experimental work.

\subsubsection{Implementation of \BadMatch}
\label{exp:SM}
In this section, MedCLIP is fine-tuned through the \textit{semantic matching loss} as prescribed by Wang et al.\cite{wang2022medclip}, represented in Eq~\eqref{eq:semLoss}. We only sample a small amount of the image data to flip their labels and poison the image following one of the trigger strategies in Sec.~\ref{exp:poi_data}, without changing any vanilla MedCLIP training algorithm. As detailed in Sec.~\ref{method:semanticM}, such small amount of label-flipping will construct the $SM_\texttt{poi}$, which further interferes with the $PM$ and then the downstream performance of MedCLIP.

By default, our experiments utilize a batch size of 32 unless specified otherwise. Our configuration involves two particularly sensitive hyper-parameters: the fine-tuned iterations and the proportion of data designated for poisoning. One might intuitively surmise that increasing the proportion of poisoned data or extending the fine-tuned iterations would enhance the backdoor attack.

To systematically ascertain the optimal boundaries for these hyper-parameters, our approach begins with a fixed training iteration and an initial, modest proportion of poisoned data. We then incrementally elevate the proportion of poisoned samples, monitoring the effectiveness of the poisoning. Upon achieving a satisfactory backdoor effect, we iteratively decrease the fine-tuned iteration to determine its influence on the outcome. A comprehensive breakdown of the diverse configurations tested can be found in Appendix~\ref{app:sm-setting}.

\subsubsection{Implementation of \ours}
\label{exp:baddist}
Revisiting Sec.\ref{method:baddist}, the objective of \ours is to introduce a discernible distance between the embeddings of clean and poisoned data, while ensuring the embeddings of the clean data remain consistent. We frame this as an optimization problem and present the corresponding objective function in Eq~\eqref{eq:baddist}.

% For clarity, Alg~\ref{alg:baddist} offers a step-by-step pseudocode detailing how \ours is implemented.

In all experiments detailed in our main paper, we optimized MedCLIP using the \ours. We employed the SGD optimizer with a learning rate set to $1 \times 10^{-4}$ and train 200 epochs. The choice of the learning rate was determined after a series of preliminary tests, ensuring optimal convergence for our model. Furthermore, \ours introduces two trade-off hyper-parameters, $\lambda_1$ and $\lambda_2$, that balance the \ours loss. Through meticulous hyper-parameter tuning, ensuring the best trade-off between \textit{effectiveness} and \textit{utility}, we set $\lambda_1=5$ and $\lambda_2=1$ consistently across all experimental setups.

For \ours-assisted \BadMatch, we run \ours first and then apply \BadMatch, with the same configuration as in Sec.~\ref{exp:imple_detail}.

\subsection{Results and Analysis}

\begin{table*}[t]
\centering
\caption{Performance of backdoor attack in COVIDX dataset under different settings. Each experiment was conducted five times, with both the average and standard deviation (indicated within brackets) reported.}
\label{tab:covidx}
\resizebox{\linewidth}{!}{%
\begin{tabular}{clccccccc}
\toprule
\multirow{2}{*}{Trigger} & \multirow{2}{*}{Strategy} & \multicolumn{3}{c}{ResNet} & \multicolumn{3}{c}{ViT} \\ 
\cmidrule(l){3-5} \cmidrule(l){6-8}
 & & BA $\uparrow$ & BSR  $\uparrow$ & Avg $\uparrow$ & BA  $\uparrow$ & BSR  $\uparrow$ & Avg $\uparrow$\\ 
\midrule
\multirow{4}{*}{Patch} &
BadEncoder & 0.7557 (0.03) & 0.3624 (0.10) & 0.5591 & 0.7796 (0.01) & 0.7083 (0.01) & 0.7440 \\

& BadEncoder-assisted \BadMatch & 0.7966 (0.00) & 0.9946 (0.10) & 0.8956 & 0.7999 (0.00) & 0.9235 (0.01) & 0.8617\\

& \BadMatch & 0.7519 (0.00) & 0.9824 (0.01) & 0.8672 & 0.7857 (0.00) & 0.9802 (0.00) & 0.8830 \\

& \ours-assisted \BadMatch & 0.7967 (0.00) & 0.9949 (0.00) & \textbf{0.8958} & 0.7925 (0.00) & 0.9939 (0.00) & \textbf{0.8932}\\ 

\midrule

\multirow{4}{*}{Fourier} &
BadEncoder & 0.7317 (0.03) & 0.9996 (0.00) & \textbf{0.8657} & 0.7178 (0.00) & 0.9997 (0.00) & 0.8588 \\
& BadEncoder-assisted \BadMatch & 0.7327 (0.02) & 0.9546 (0.01) & 0.8437 & 0.6868 (0.01) & 0.9997 (0.00) & 0.8433\\

& \BadMatch & 0.7370 (0.02) & 0.9095 (0.02) & 0.8233 & 0.6382 (0.01) & 1.0000 (0.00) & 0.8191 \\

& \ours-assisted \BadMatch & 0.7352 (0.02) & 0.9931 (0.00) & 0.8642 & 0.7743 (0.01) & 1.0000 (0.00) & \textbf{0.8872} \\ 

\bottomrule
\end{tabular}
}
\end{table*}

\begin{table*}[b]
\centering
\caption{Performance of backdoor attack in RSNA dataset under different settings. Each experiment was conducted five times, with both the average and standard deviation (indicated within brackets) reported.}
\label{tab:rsna}
\resizebox{\textwidth}{!}{%
\begin{tabular}{clccccccc}
\toprule
\multirow{2}{*}{Trigger} & \multirow{2}{*}{Strategy} & \multicolumn{3}{c}{ResNet} & \multicolumn{3}{c}{ViT} \\ 
\cmidrule(l){3-5} \cmidrule(l){6-8}
 & & BA $\uparrow$ & BSR  $\uparrow$ & Avg $\uparrow$ & BA  $\uparrow$ & BSR  $\uparrow$ & Avg $\uparrow$\\ 
\midrule
\multirow{4}{*}{Patch} &
BadEncoder & 0.6942 (0.08) & 0.5731 (0.20) & 0.6337 & 0.7663 (0.01) & 0.4324 (0.06) & 0.5994 \\

& BadEncoder-assisted \BadMatch & 0.7616 (0.08) & 0.5204 (0.20) & 0.6410 & 0.7557 (0.01) & 0.5260 (0.00) & \textbf{0.6409} \\

& \BadMatch & 0.7405 (0.05) & 0.5081 (0.09) & 0.6243 & 0.7589 (0.00) & 0.3438 (0.02) & 0.5514 \\

& \ours-assisted \BadMatch & 0.7064 (0.03) & 0.6321 (0.07) & \textbf{0.6693} & 0.7549 (0.01) & 0.4729 (0.05) & 0.6139 \\ 

\midrule

\multirow{4}{*}{Fourier} &
BadEncoder & 0.6592 (0.04) & 1.0000 (0.00) & 0.8296 & 0.7178 (0.00) & 0.9970 (0.00) & 0.8574 \\

& BadEncoder-assisted \BadMatch & 0.6405 (0.04) & 0.9708 (0.00) & 0.8057 & 0.7065 (0.01) & 0.9989 (0.00) & 0.8527 \\

& \BadMatch & 0.6647 (0.03) & 0.9158 (0.04) & 0.6903 & 0.6683 (0.02) & 0.9987 (0.00) & 0.8335 \\

& \ours-assisted \BadMatch & 0.7244 (0.01) & 0.9923 (0.00) & \textbf{0.8584} & 0.7244 (0.01) & 1.0000 (0.00) & \textbf{0.8622} \\ 

\bottomrule
\end{tabular}
}
\end{table*}

Table~\ref{tab:covidx} details the results of the targeted backdoor attack on the COVIDX dataset, whereas Table~\ref{tab:rsna} displays the findings for the RSNA dataset. We employ the evaluation metrics designed for targeted attacks, namely, BA and BSR, as detailed in Sec.~\ref{method:goal}. Notably, there exists an intrinsic trade-off between BA and BSR. A particularly \textit{effective} backdoor attack, for instance, might gain a high BSR. However, this often leads to a compromise in \textit{utility}, manifested as a reduced BA. To offer a holistic view of the backdoor attack's efficacy, the final column for each experimental group provides an average of both BA and BSR.

\subsubsection{Analysis the effect of \BadMatch} 
\label{exp:result-unpair}

In this section, we assess the performance of the backdoor introduced via \BadMatch, wherein a subset of images has been poisoned with flipping labels. The outcomes arising purely from \BadMatch are delineated in the third row of both tables.

Focusing on the COVIDX dataset, as depicted in Table~\ref{tab:covidx}, both ResNet and ViT yield a BSR exceeding 90\% across all triggers. Notably, the Fourier attack under ViT even reaches a 100\% BSR. Turning our attention to the RSNA dataset in Table~\ref{tab:rsna}, the Fourier trigger achieves a BSR surpassing 90\%. It's important to highlight that the poisoned data for all Fourier strategies constitute a mere 0.5\% of the fine-tuned subset, as evidenced in Appendix~\ref{app:sm-setting}. This implies that even with a minuscule percentage of mislabeled data, the unpaired training can amplify its effect, resulting in a backdoor attack effectiveness of over 90\%.

As introduced in Sec.~\ref{method:semanticM}, the inherent nature of the unpaired training strategy is to amplify the dataset by pairing congruent data from varying modalities, e.g., vision and language in our case. Consequently, even a slight fraction of incorrectly paired data can magnify the detrimental aspects of this approach. 
This observed phenomenon underscores the crucial importance of thorough data validation prior to employing such a strategy.

\subsubsection{Analysis the effect of BadEncoder}
The first row of Table~\ref{tab:covidx} and Table~\ref{tab:rsna} presents the performance of BadEncoder on MedCLIP.

With patch-based triggers, BadEncoder's effectiveness is limited in terms of BSR. Specifically, for the COVIDX dataset, the average BSR is a modest 53\% across both neural architectures. This may be attributed to our patching strategy, where patches are concealed within their surrounding environment, making it challenging for the neural network to discern them.

In contrast, when employing Fourier-based triggers, BadEncoder achieves a commendable BSR of approximately 100\%, while preserving the BA. However, the average values for BA and BSR do not rank as the highest in the table.

\subsubsection{Analysis the effect of \ours-assisted and BadEncoder-assited \BadMatch} 
\label{exp:result-target}
In this part, we analyze the results for BadEncoder-assisted and \ours-assisted \BadMatch. The respective outcomes are detailed in the second and fourth rows of Table~\ref{tab:covidx} and Table~\ref{tab:rsna}.

For Fourier-based backdoor attacks, we consistently observe an elevated BSR across all test groups. The average BSR peaks at a commendable 98.87\% for all datasets and neural architectures. This performance, in terms of \textit{effectiveness}, is notably superior when juxtaposed with the outcomes of \BadMatch when used singly. The pronounced efficacy stems from the synergistic interaction of the combined strategies with the FM's embedding, inherently altering the predictive matrix, $PM$, to streamline it for the designated backdoor attack.

Simultaneously, in certain test cases, the \ours-assisted \BadMatch reaches high BA values. For instance, within the Fourier-ResNet-RSNA group (Table~\ref{tab:rsna}), \ours-assisted \BadMatch achieves a BA of 72.44\%, marking a 13\% improvement compared to its BadEncoder-assisted counterpart in the identical group. Similarly, in the Fourier-ViT-COVIDX group, \ours-assisted \BadMatch attains a 77.43\% BA, a notable improvement from the 71.78\% of the runner-up.

Turning our attention to patch-based backdoor attacks, the average BSR for the COVID dataset stands at 86.75\%, with \ours-assisted \BadMatch eclipsing 99\% for both ResNet and ViT. All BA scores surpass 75\%, with those aided by \ours and BadEncoder hovering around 79\%. Hence, this backdoor demonstrates commendable \textit{effectiveness} and \textit{utility} on the COVIDX dataset. In the RSNA dataset's context, the mean BSR is a modest 50\%, with variations based on architecture and assisting method. Given the depth to which patch triggers are concealed within their surroundings, a subdued BSR, in comparison to other groups, is not surprising.

Lastly, inspecting the average column, \ours-assisted \BadMatch emerges as the top performer in six out of eight test groups. In the remaining two, \ours maintains competitive scores, closely trailing the best performers. This highlights the robustness and efficacy of \ours across different configurations.

Our analysis reveals that the \ours-assisted \BadMatch strikes a fine balance between BA and BSR, as corroborated by the aggregated scores. A potential rationale behind this observation is that BadEncoder is inherently anchored to the embeddings of the poisoned samples in the initial encoder. When there exists a high similarity between clean and poisoned image embeddings, BadEncoder's efficacy in modifying the $PM$ wanes, leading to suboptimal attacks in terms of both \textit{effectiveness} and \textit{utility}. This challenge is more pronounced when poisoned images bear a close resemblance to their clean counterparts, as our patch experiments demonstrate. In such scenarios, \ours stands out, owing to its strategy of distinguishing embeddings based on their relative spatial positioning.

\subsubsection{Untargeted attack with \ours} 
\label{exp:result-untarget} 
\begin{figure}
\subfloat[]{\includegraphics[width=0.5\linewidth]{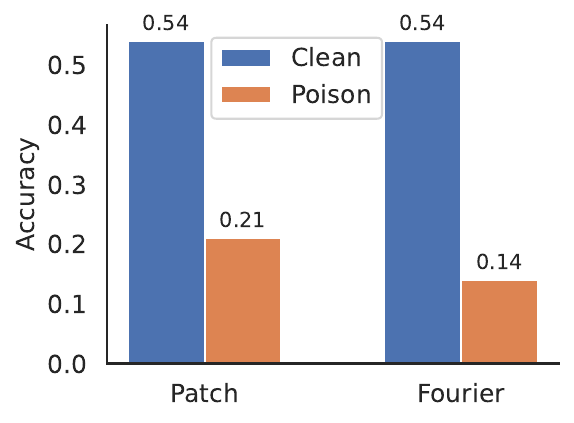}}
\subfloat[]{\includegraphics[width=0.5\linewidth]{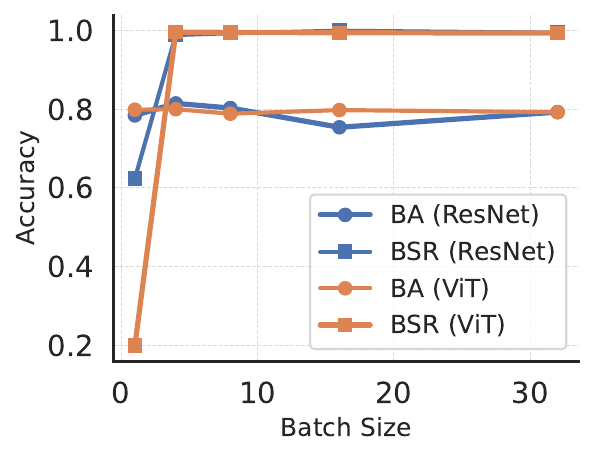}}
    \caption{(a) Classification accuracy for ViT-based MedCLIP under untargeted attacks using Patch and Fourier-based trigger strategies. (b) \ours-assisted \BadMatch under different batch sizes for patch-based backdoor on COVIDX.}
    \label{fig:untargeted}
\end{figure}

In this section, we focus exclusively on \ours to evaluate its impact on untargeted attacks using the MIMIC test set, designed for 10-class classifications. Our experimental setup mirrors that of Sec.\ref{exp:baddist}, with the sole difference being the exclusive application of \ours to MedCLIP. Notably, alternative strategies, such as BadEncoder\cite{jia2022badencoder}, are not amenable to untargeted attacks. We conduct our experiments using the ViT backbone, and our findings are graphically represented in Fig~\ref{fig:untargeted}.

For the MIMIC dataset, the baseline classification accuracy of a pre-trained MedCLIP is recorded at 55\%. As delineated in Fig~\ref{fig:untargeted}, the accuracy for clean data remains consistent at approximately 54\%, irrespective of whether patch-based or Fourier-based trigger strategies are deployed. In contrast, for poisoned data, the accuracy nosedives to around 21\% with the patch-based strategy—signifying a 63\% decrement. The Fourier-based triggers fare even more poorly, with accuracy plummeting to a mere 13\%, akin to making random predictions.

These observations further underscore the potency of \ours in manipulating the predictive matrix, $PM$, reinforcing its capability to effectively compromise contrastive model performance.

% \begin{figure}
% \includegraphics[width=\linewidth]{figures/untarget.pdf}
%     \caption{Classification accuracy for ViT-based MedCLIP under untargeted attacks using Patch and Fourier-based trigger strategies.}
%     \label{fig:untargeted}
% \end{figure}

\subsubsection{Assessing \BadMatch's stability across varied batch sizes}
\label{exp:result-batch}
In \BadMatch, poisoned images are aligned with target sentences within a ${SM}_{\texttt{poi}}$ in the training batch. This raises the question of \textit{how batch size, $|N|$, might influence this matching}. For instance, in notably small batch sizes, all images could belong to a single class, precluding any match with sentences from the target class. In this subsection, we conduct an ablation study spanning multiple batch sizes, demonstrating that \ours-assisted \BadMatch exhibits consistent performance across varied batch dimensions.

% Fig~\ref{fig:batch} 
Fig~\ref{fig:untargeted} (b) illustrates the BA and BSR trends across different batch sizes for both the ResNet and ViT architectures on the COVIDX dataset. Again, except for the batch size, $|N|$, everything stays the same as the experiments in Table~\ref{tab:covidx}. we observe the following: At $|N|=1$, the image exclusively aligns with its corresponding text, leading to a notably diminished BSR. As the batch size incrementally expands to four, the BSR approaches its peak value, nearly 1. After this, the BA and BSR stays the same with the increase of the batch size. Concurrently, the BA remains consistent across both neural structures, reinforcing the stability of our proposed method even at reduced batch dimensions.

% \begin{figure}
% \includegraphics[width=\linewidth]{figures/batch_study.pdf}
%     \caption{\ours-assisted \BadMatch under different batch sizes for patch-based backdoor on COVIDX.}
%     \label{fig:batch}
% \end{figure}

\subsection{Defenses}
We examine two defense strategies introduced in~\ref{pre:defense}: \textit{empirical} and \textit{certified} defenses. Since these defense mechanisms are primarily designed for patch-based triggers, we evaluate it on \ours-assisted \BadMatch on COVIDX dataset using Resnet.

\textbf{Empirical Defense}
seeks to mitigate or identify the effects of backdoor attacks using empirical analyses. Data augmentation is an exemplary strategy, particularly when backdoor triggers often manifest in specific areas of data~\cite{qiu2021deepsweep, li2020rethinking}. Evaluating its effectiveness, we first examine its impact on zero-shot classification, employing Optical Distortion and Stochastic Affine Transformation during the inference phase. The result indicates a modest reduction in the BSR from $0.9977$ to $0.9813$, underscoring the \textit{limited} success of this defensive approach against our attack. Following this, we implement the empirical defense. MNTD~\cite{xu2021detecting} suggests starting by creating a series of backdoored models, then training meta classifiers on both the uncorrupted and corrupted models to jointly assess the integrity of it. Adapting this methodology to our scenario, we retain the parameters of the original vision encoder and train 100 pairs of both clean and corrupted linear classifiers, considering them as our meta-training dataset. For testing, 20 pairs of linear classifiers are trained on both the clean encoder and our altered version, using the previously trained meta-classifiers for detection purposes. The average accuracy across meta classifiers in detecting our attack is 0.67, highlighting the partial efficacy of MNTD against our approach. Notably, while the authors of the original work argue that a dataset of 64 training pairs is adequate for detecting backdoor attacks, we expanded this to 100 pairs for a more comprehensive assessment.

\textbf{Certified Defense}
emphasizes the identification of compromised trained models based on the intrinsic attributes of backdoor attacks. Our attack efficacy is measured against an established provable defense, PatchGuard~\cite{xiang2020patchguard}. This defense unequivocally demonstrates that backdoor attacks can be effectively countered in neural networks with smaller receptive fields, provided the trigger size remains below a certain threshold. By systematically masking dubious features, it can yield a minimum accuracy, dubbed the \textit{certified accuracy}. Following this methodology, we freeze the vision encoder and train a linear head. Leveraging the publicly available implementation of PatchGuard with a patch size set to $32$, we test our patch-ResNet-COVIDX group. The outcomes indicate a certified accuracy of $0.085$, and a \textit{clean accuracy} post robust masking of $0.369$. This suggests that PatchGuard doesn't provide an adequate defense against the attack.

%% file: sections/conclusion.tex
\section{Conclusion}
In this paper, we explored the vulnerabilities inherent to \textit{unpaired} matching, termed BadMatch. Our investigation reveals that a modest 5\% misalignment in image labeling can lead to an augmented effect in the process of poisoned inputs being erroneously matched with incorrect text labels. Astonishingly, this slight misalignment, when leveraged through the unpaired training strategy, can culminate in a backdoor attack efficacy nearing 99\%. The introduction of \ours-assisted \BadMatch further delineates the chasm between the embeddings of clean and poisoned data, compelling the FM to generate distinct features for each, which subsequently impacts various downstream tasks.

In essence, our research sheds light on the resilience of backdoor attacks across a broad spectrum: spanning different model architectures, datasets, poisoning techniques, and batch sizes inherent to the FM supply chain. This work stands as a pioneering effort in identifying potential pitfalls in unpaired training and the susceptibilities of pre-trained contrastive FMs in the face of backdoor adversaries. Our findings underscore the critical importance of meticulous data validation in unpaired training setups and emphasize the need for stringent model validation processes within the model supply chain.

%% file: sections/acknowledgement.tex
\section{Acknowledgment}
This work is supported in part by the Natural Sciences
and Engineering Research Council of Canada (NSERC), Public Safety Canada, CIFAR Catalyst Grant, Compute Canada Research Platform, and Microsoft Accelerate Foundation Models Research Program.

%% file: sections/appendix.tex
\section{Notation Table}
\label{app:appendix_notation}
Table~\ref{tab:notation} presents a comprehensive list of mathematical notations utilized throughout this paper, accompanied by their respective descriptions.

\begin{table}[H]
\caption{}
\centering
\resizebox{\columnwidth}{!}{
\begin{tabular}{cl}
\toprule
Notations & Description \\ 
\hline \\[-1.8ex]
$b$, $b'$ & poison data embedding from clean model and backdoor model individually\\

$c$, $c'$ & clean data embedding from clean model and backdoor model individually\\

$h$ & hidden dimension for encoders \\

$v$& text embedding \\

$t$ & image embedding \\
\vspace{2mm}
$l$ & image or text label \\

$x_{img}$, $x_{txt}$ & image and text data \\
\vspace{2mm}
$x_{trigger}$ & trigger generated by one of the poison strategies in Sec.~\ref{exp:poi_data} \\
\vspace{2mm}
$y_{img}^{target}$ & target image label \\

$\tau$ & temperature in contrastive loss \\

$\lambda_1$ & coefficient for clean data \\ 

$\lambda_2$ & coefficient for poison data \\ 

$P$ & set of poisoned data \\

$N$ & set of batch indexes \\

$K$ & set of label vector indexes \\

$SM$ & sematic (similarity) matrix \\

$SM_\texttt{poi}$ & poisoned sematic (similarity) matrix \\

$PM$ & predictive (similarity) matrix \\

$f_{\theta}$, $\tilde{f}_{\theta}$ & clean model and poisoned model \\

$\mathcal{L_{\rm MedCLIP}}$ & semantic matching loss used by original MedCLIP \\ 

$\texttt{sim}(c,d)$ & the cosine similarity between vector $c$ and $d$ \\
\bottomrule
\end{tabular}
}
\label{tab:notation}
\end{table}

\section{Detailed Configuration}
\label{app:sm-setting}

Table~\ref{tab:detailed_config} delineates the range of examined poisoned proportions $p$ in Alg.~\ref{alg:semantic_matrix} and the iterations fine-tuned. The final configurations of all hyper-parameters are highlighted in \textbf{bold}. As detailed in Section~\ref{exp:SM}, the process commences at iteration 4000. The proportion of poisoned data is incrementally raised from 0.05 until the desired backdoor effect is observed. Subsequently, the training iteration is methodically reduced to discern the boundary conditions.

\begin{table}[H]
\centering
\caption{}
\resizebox{1\linewidth}{!}{
\begin{tabular}{@{}ccc@{}}
\toprule
Setting & Range Examined (poison proportion, iteration) \\ %& Final Specification  \\
\midrule
Covid-Patch-ResNet & \makecell[c]{[(0.05, 4000), (0.1, 4000), (0.15, 4000),\\ \textbf{(0.2, 4000)},(0.2, 3500),(0.2, 3000)]}  \\
\midrule
Covid-Patch-ViT & \makecell[c]{[(0.05, 4000), (0.1, 4000), (0.15, 4000), \\ (0.2, 4000),(0.2, 3000),(0.2, 3000),(0.2, 2500), \textbf{(0.2, 2000)}, (0.2,1500)}  \\
\midrule
RSNA-Patch-ResNet  & \makecell[c]{[(0.05, 4000),\\ (0.1. 4000), (0.1, 3500),(0.1, 3000), (0.1, 2500), (0.1, 2000), \\ \textbf{(0.1, 1500)}, (0.1, 1000)]} \\
\midrule
RSNA-Patch-ViT & \makecell[c]{[(0.05, 4000), (0.05, 3500),(0.05, 3000), (0.05, 2500), (0.05, 2000), \\(0.05, 1500), \textbf{(0.05, 1000)}, (0.05, 500)]}  \\
\midrule
Covid-Fourier-ResNet & \makecell[c]{[(0.05, 4000), (0.05, 3500), (0.05, 3000),(0.05, 2500), \\ (0.05, 2000),(0.05, 1500), (0.05,1000), \textbf{(0.05,500)}]} \\
\midrule
Covid-Fourier-ViT & \makecell[c]{[(0.05, 4000), (0.05, 3500), (0.05, 3000),\\ (0.05, 2500), \textbf{(0.05, 2000)}, (0.05, 1500)]} \\
\midrule
RSNA-Fourier-ResNet  & \makecell[c]{[(0.05, 4000), (0.05, 3500), (0.05, 3000),(0.05, 2500),\\ (0.05, 2000),(0.05, 1500), (0.05,1000),\textbf{(0.05,500})]}\\
\midrule
RSNA-Fourier-ViT & \makecell[c]{[(0.05, 4000), (0.05, 3500), (0.05, 3000),\\ (0.05, 2500), \textbf{(0.05, 2000)}, (0.05, 1500)]} \\
\bottomrule
\end{tabular}
}
\label{tab:detailed_config}
\end{table}